\documentclass[conference]{IEEEtran}
\IEEEoverridecommandlockouts
\usepackage{amsmath,amssymb,amsfonts}
\DeclareMathOperator*{\argmin}{\arg\!\min}
\usepackage{algorithmic}
\usepackage{graphicx}
\usepackage{subcaption}
\usepackage{textcomp}
\usepackage{multirow}
\usepackage{tikz}
\usepackage{pgfplots}
\pgfplotsset{compat=1.18}
\usepackage{xcolor}
\usepackage[acronym,nohypertypes={acronym,notation}]{glossaries}
\usepackage{float}
\newacronym{mdp}{MDP}{Markov Decision Process}
\newacronym{swarmdp}{SwarMDP}{Swarm Markov Decision Process}
\newacronym{pomdp}{POMDP}{Partially Observable Markov Decision Process}
\newacronym{mbrl}{MBRL}{Model-Based Reinforcement Learning}
\newacronym{gnn}{GNN}{Graph Neural Network}
\newacronym{mpnn}{MPNN}{Message Passing Neural Network}
\newacronym{mpn}{MPN}{Message Passing Network}
\newacronym{mlp}{MLP}{Multilayer Perceptron}
\newacronym{cnn}{CNN}{Convolutional Neural Network}
\newacronym{rl}{RL}{Reinforcement Learning}
\newacronym{srl}{Swarm RL}{Swarm Reinforcement Learning}
\newacronym{nn}{NN}{Neural Network}
\newacronym{ppo}{PPO}{Proximal Policy Optimization}
\newacronym{osi}{OSI}{Open Systems Interconnection}
\newacronym{iqm}{IQM}{Interquartile Mean}
\newacronym{dqn}{DQN}{Deep Q-Network}
\newacronym{rp}{RP}{Routing Protocol}
\newacronym{ml}{ML}{Machine Learning}
\newacronym{te}{TE}{Traffic Engineering}
\newacronym{ro}{RO}{Routing Optimization}
\newacronym{sdn}{SDN}{Software-Defined Networking}
\newacronym{kdn}{KDN}{Knowledge-Defined Networking}
\newacronym{sr}{SR}{Segment Routing}
\newacronym[longplural={Traffic Matrices}]{tm}{TM}{Traffic Matrix}
\newacronym{ls}{LS}{Local Search}
\newacronym{marl}{MARL}{Multi-Agent Reinforcement Learning}
\newacronym{lu}{LU}{Link Utilization}
\newacronym{idcwan}{Inter-DC WAN}{Inter-Datacenter Wide Area Network}
\newacronym{p2p}{P2P}{Point-to-Point}
\newacronym{ba}{BA}{Barab\'{a}si-Albert}
\newacronym{er}{ER}{Erd\H{o}s-R\'{e}nyi}
\newacronym{ws}{WS}{Watts-Strogatz}
\newacronym{ospf}{OSPF}{Open Shortest-Path First}
\newacronym{eigrp}{EIGRP}{Enhanced Interior Gateway Routing Protocol}
\newacronym{spf}{SPF}{Shortest-Path First}
\newacronym{ecmp}{ECMP}{Equal-Cost Multipath}
\newacronym{mpls}{MPLS}{Multiprotcol Label Switching}
\newacronym{qos}{QoS}{quality of service}
\newacronym{int}{INT}{In-Band Network Telemetry}
\newacronym{ip}{IP}{Internet Protocol}
\newacronym{sack}{SACK}{Selective Acknowledgement}
\newacronym{udp}{UDP}{User Datagram Protocol}
\newacronym{tcp}{TCP}{Transmission Control Protocol}
\newacronym{aoi}{AoI}{Age of Information}
\newacronym{apsp}{APSP}{All Pairs Shortest Paths}
\newacronym{mptcp}{MPTCP}{Multipath TCP}
\newacronym{bc}{BC}{Behavioral Cloning}
\newacronym{il}{IL}{Imitation Learning}
\newacronym{tmlr}{TMLR}{Transactions on Machine Learning Research}  
\usepackage[
  backend=biber,
  style=numeric,
  natbib=true,
  url=false,
  doi=false,
  isbn=false,
  eprint=false
]{biblatex}

\begin{document}

\title{Can Neural Networks Provide Latent Embeddings for Telemetry-Aware Greedy Routing?
\thanks{This research work was carried out as part of the project “Apeiro Reference Architecture”, in short ApeiroRA, under the funding ID 13IPC007, financed by the European Union – NextGenerationEU and funded by the German Federal Ministry for Economic Affairs and Energy.
\par
Correspondence to: Andreas Boltres \textless andreas.boltres@partner.kit.edu\textgreater}
}

\author{%
  \IEEEauthorblockN{%
    Andreas Boltres\IEEEauthorrefmark{1}\IEEEauthorrefmark{2},
    Niklas Freymuth\IEEEauthorrefmark{1} and
    Gerhard Neumann\IEEEauthorrefmark{1}
  }%
  \IEEEauthorblockA{\IEEEauthorrefmark{1} Autonomous Learning Robots, Karlsruhe Institute of Technology, Germany}%
  \IEEEauthorblockA{\IEEEauthorrefmark{2} SAP SE, Walldorf, Germany}%
}

\maketitle

\begin{abstract}
Telemetry-Aware routing promises to increase efficacy and responsiveness to traffic surges in computer networks. Recent research leverages Machine Learning to deal with the complex dependency between network state and routing, but sacrifices explainability of routing decisions due to the black-box nature of the proposed neural routing modules. We propose \emph{Placer}, a novel algorithm using Message Passing Networks to transform network states into latent node embeddings. These embeddings facilitate quick greedy next-hop routing without directly solving the all-pairs shortest paths problem, and let us visualize how certain network events shape routing decisions.
\end{abstract}


\vspace{-0.1cm}

\section{Introduction}
In computer networks, routing denotes the task of selecting paths along which data packets are forwarded. A good routing mechanism is essential for an efficient use of the given infrastructure. Routing optimization has been a long-standing subject of study because traffic often fluctuates with unpredictable patterns~\citep{alizadehCONGADistributedCongestionaware2014, wendell2011going}, influencing the optimal routing solution. Recent work has suggested that routing decisions be based on live telemetry data~\citep{turkovicFastNetworkCongestion2018, xieDelayguaranteedRoutingMechanism2022} and to re-optimize routes within milliseconds~\citep{gayExpectUnexpectedSubsecond2017, guiRedTEMitigatingSubsecond2024} to improve routing responsiveness. Yet, for existing methods, the complex, non-linear dependency between high-dimensional network state and suitable routing becomes challenging to navigate, especially as networks increase in scale~\citep{xuLinkStateRoutingHopbyHop2011}. Leveraging \gls{ml} for network routing has been a subject of study for decades~\citep{boyanPacketRoutingDynamically1993} that has seen differing problem formulations, network types, use cases and goals. Current research derives splitting ratios for precomputed sets of shortest paths~\citep{xiaoLeveragingDeepReinforcement2021, xuTealLearningAcceleratedOptimization2023} from traffic aggregated over minutes to hours, and infers routes directly for individual packets~\citep{maiPacketRoutingGraph2021} and packet flows~\citep{zhangFlowletLevelRoutingOptimization2023}. Related work constructs node-level network embeddings that facilitate greedy routing by minimizing the remaining distance to the destination in the embedding space~\citep{kuhnGeometricAdhocRouting2003, kleinbergGeographicRoutingUsing2007, blasiusHyperbolicEmbeddingsNearOptimal2020a}. With a suitable network embedding, these approaches promise competitive routing without the computational cost of computing shortest paths. 
\begin{figure}[thbp]
    \centerline{\includegraphics[scale=0.0435]{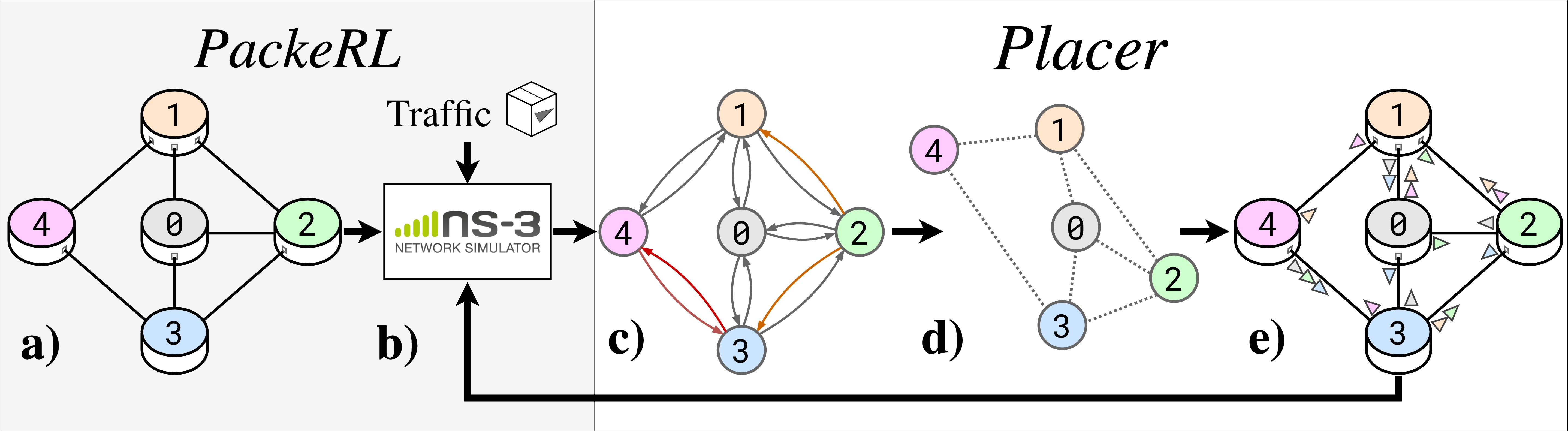}}
    \caption{
        \textbf{a)}: Five-node network topology \emph{mini-5} used in our experiments. \textbf{b, c)}: \emph{PackeRL}'s simulation backend turns network topology, incoming traffic and routing actions into an attributed state graph. \textbf{d, e)}: \emph{Placer} obtains latent node embeddings using its Message Passing Network, then greedily converts these into next-hop selections.
    }
    \label{vis}
    \vspace{-0.4cm}
\end{figure}
While existing work uses \gls{ml} to obtain node-level representations from IP network information~\citep{liDeepLearningIP2018}, none have yet used \gls{ml} to turn network state graphs containing telemetry data into node embeddings for greedy routing.

We propose to connect both aspects, providing telemetry-aware embeddings for IP routing in milliseconds. Our approach, \emph{Placer}, uses \glspl{mpn}~\citep{gilmer2017neural} to obtain latent Euclidean node embeddings from telemetry-infused network state graphs. To obtain informative network states within realistic network environments, we use \emph{PackeRL}~\citep{boltres2024learning}, a training and evaluation framework for sub-second telemetry-aware routing built on top of the packet-level network simulator \emph{ns-3}~\citep{hendersonNetworkSimulationsNs3} and its capabilities for in-band network telemetry~\citep{kim2015band}.

\section{Problem Setting}

We follow~\citet{boltres2024learning} and view routing in IP networks as a Markov Decision Process $\langle \mathcal{S}, \mathcal{A}, \mathcal{T}, r \rangle$. States $S_t=(V_t,E_t,\mathbb{X}_{V_t,t}, \mathbb{X}_{E_t,t}, \mathbf{x}_{u,t}) \in \mathcal{S}$ are attributed directed graphs that hold topology information like link data rate or packet buffer size, as well as telemetry data obtained via \mbox{\emph{ns-3}'s} \emph{FlowMonitor}~\citep{carneiroFlowMonitorNetworkMonitoring2009}. The action \mbox{$\mathbf{a}_t = \{(u, z) \mapsto v \mid u, v, z \in V_t, v \in \mathcal{N}_u\} \in \mathcal{A}$} selects next-hop neighbors per routing node $u \in V_t$ for each possible destination node $z \in V_t$. The transition function \mbox{$\mathcal{T}: \mathcal{S} \times \mathcal{A} \rightarrow \mathcal{S}$} evolves the current network state using the latest routing actions. In \emph{PackeRL}, this means installing the actions and invoking \mbox{\emph{ns-3}'s} simulator for a predetermined duration. The global reward function $r: \mathcal{S} \times \mathcal{A} \rightarrow \mathbb{R}$ assesses a routing action in the given network state. We use the global goodput, measured as the sum of MB received per step across all nodes of the topology. We train a policy $\pi: \mathcal{S} \times \mathcal{A} \rightarrow [ 0, 1 ]$ to maximize the expected discounted cumulative future reward
$J_t:=\mathbb{E}_{\pi(\mathbf{a}|\mathbf{s})}\left[\sum_{k=0}^{\infty}\gamma^k r(\mathbf{s}_{t+k},\mathbf{a}_{t+k})\right]$~\citep{Sutton1998}, and thus the long-term global goodput.

\newpage

\section{From Node Embeddings to Greedy Routing}

\emph{Placer} consists of two phases: First, it uses a four-layer \gls{mpn} with a hidden dimension of $32$ to obtain latent node embeddings $\mathbf{x}_i \in {\mathbb{R}^d}$ of configurable output dimensionality $d$ from the current network state $S_t$. Then, it decomposes each embedding $\mathbf{x}_i$ into its radius $r_i=||\mathbf{x}_i||$ and unit direction $u_i = \frac{\mathbf{x}_i}{r_i}$, and computes pairwise squared distances $\Delta^2 \in [0, 4)^{|V|\times |V|}$ as $\Delta_{ij}^2=\rho_i^2+\rho_j^2-2\rho_i\rho_j (u_i^\top u_j)$, using the law of cosines in polar form with softly constrained radii $\rho_i = \tanh(r_i) \in [0,1)$. For any given pair of routing node~$u$ and destination node $z$, greedy routing in the embedding space chooses the next-hop node $v = \argmin_{v' \in \mathcal{N}_{u}} \Delta(v',z)$ that minimizes the remaining distance to $z$.
\par
To train \emph{Placer}, we use the \gls{ppo}~\citep{schulman2017proximal} implementation and hyperparameters of~\citet{boltres2024learning}. We implement \textit{Boltzmann exploration}~\citep{kaelblingReinforcementLearningSurvey1996} by treating the values of~$\Delta^2$ as negative logits for $|V|^2$ next-hop sampling distributions, one per pair of $(u, z)$.

\section{Experiments and Results}

We use the episodic experiment setup of~\citet{boltres2024learning} and a five-node network topology shown in Figure~\ref{vis}\textbf{a)} which we call \emph{mini-5}. 
We simulate $H=400$ steps of $5$~ms each per episode, injecting synthetic TCP and UDP flows which, in total, exceed the network's capacity. Flow sizes and arrival times are generated akin to~\citet{boltres2024learning}, following distributions found in real-world data centers~\citep{benson2010network}. 
For \emph{Placer} (ablated with $d = i \in  \{1, 2, 32\}$ and written as \emph{Placer}$_{d=i}$) and the baseline algorithm \emph{M-Slim}~\citep{boltres2024learning}, we use $40$ \gls{ppo} iterations of $16$ episodes, each providing $4$ unique traffic sequences repeated $4$ times. We report the interquartile mean over $8$ random seeds~\citep{agarwalDeepReinforcementLearning2021}, with performance averaged over 30 unseen traffic sequences. Lastly, we use the routing scheme of Cisco's \gls{eigrp}~\citep{savageCiscoEnhancedInterior2016} as a telemetry-oblivious baseline which, by default, computes shortest paths from link data rate and delay values.

\begin{table}[htbp]
\begin{center}
\begin{tabular}{|c|c|c|c|c|c|}
\hline
 & \multirow{2}{*}{\textbf{EIGRP}}
 & \multirow{2}{*}{\textbf{\emph{M-Slim}}}
 & \multicolumn{3}{|c|}{\textbf{\textit{Placer}}} \\
\cline{4-6}
& & & $d = 1$ & $d = 2$ & $d = 32$ \\
\hline
Goodput (MB, $\uparrow$) & 209.2 & 220.45 & 236.9 & 237.0 & \textbf{237.3} \\
Avg. Delay (ms, $\downarrow$) & 9.99 & \textbf{9.59} & 9.93 & 9.92 & 9.93 \\
Dropped (\%, $\downarrow$) & 0.47 & 2.92 & 0.36 & 0.35 & \textbf{0.30} \\
\hline
Fluctuation (\%) & 0.0 & 22.80 & 0.71 & 0.58 & 0.06 \\
\hline
\end{tabular}
\caption{Mean routing performance over 30 evaluation episodes on \emph{mini-5}. Arrows denote whether lower ($\downarrow$) or higher~($\uparrow$) values are better for the respective metric.}
\label{res}
\end{center}
\end{table}

Table~\ref{res} shows the results of the aforementioned experiment for the total goodput per episode, the average packet delay, and the percentage of dropped data. Additionally, the "Fluctuation" row denotes the percentage of changes in next-hop decisions between consecutive time steps. Both \emph{M-Slim} and \emph{Placer} notably increase the episodic goodput over the telemetry-oblivious \gls{eigrp} baseline. 
\emph{M-Slim} also notably reduces packet latency, however at the expense of more packet drops. 
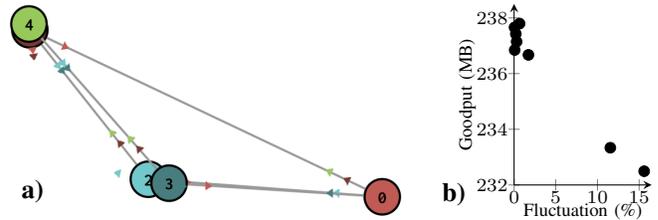
\begin{figure}[h]
    \centering
    \begin{subfigure}[t]{0.6\columnwidth}
        \centering
        \begin{tikzpicture}
            \node[anchor=south west, inner sep=0] (img) at (0,0)
                {\includegraphics[width=\linewidth]{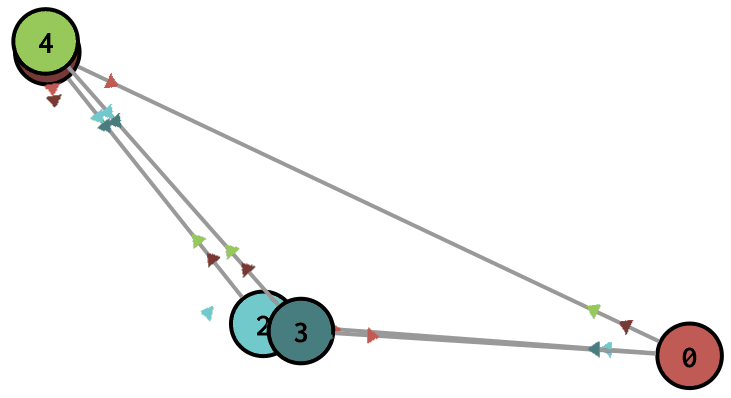}};
            \node[anchor=south west, font=\bfseries]
                at (0.02\linewidth,0.02\linewidth) {a)};
        \end{tikzpicture}
    \end{subfigure}
    \hfill
    \begin{subfigure}[t]{0.385\columnwidth}
        \centering
        \begin{tikzpicture}
            \node[anchor=south west, inner sep=0] (plot) at (0,0)
                {\begin{tikzpicture}
\begin{axis}[
    width=1\linewidth,
    height=1.17\linewidth,
    xlabel={Fluctuation (\%)},
    ylabel={Goodput (MB)},
    tick label style={font=\scriptsize},
    label style={font=\scriptsize},
    axis lines=left,
    ymin=232,
    ymax=238.5,
    xmin=0,
    xmax=16.5,
    scatter/classes={
        a={mark=*,draw=black,fill=black}
    },
]
\addplot[
    scatter,
    only marks,
    scatter src=explicit symbolic
]
table[meta=label] {
x           y           label
0.65798   237.79603   a
0.054135  237.66068   a
0.20652   237.42729   a
0.30409   237.14640   a
0.069841  236.84629   a
1.72130   236.67197   a
11.54600   233.33696   a
15.60300   232.49328   a
};
\end{axis}
\end{tikzpicture}};
            \node[anchor=south west, font=\bfseries]
                at (-0.1\linewidth,0.02\linewidth) {b)};
        \end{tikzpicture}
    \end{subfigure}
    \caption{\textbf{a)} An embedding of \emph{mini-5} produced by \textit{Placer}$_{d=2}$. \textbf{b)} Goodput vs. fluctuation of \emph{Placer}$_{d=2}$ for $8$ random seeds.}
    \label{fig2}
\end{figure}
In contrast, \emph{Placer} shows the highest goodput rates and the lowest drop percentages overall, with packet latencies comparable to \gls{eigrp}. While its performance correlates weakly with increasing $d$, interestingly, \emph{Placer}$_{d=1}$ appears to represent \emph{mini-5} well enough for competitive routing performance.

Despite the promising goodput values for \emph{Placer}, the "Fluctuation" row of Table~\ref{res} and the contents of Figure~\ref{fig2} hint at a caveat: \emph{Placer} converges to a practically telemetry-oblivious embedding of \emph{mini-5}, as indicated by the low fluctuation values. Illustrating the embeddings for \emph{Placer}$_{d=2}$, as shown in Figure~\ref{fig2}\textbf{a)}, is of limited use to human experts, as telemetry data only leads to minimal and inconsequential embedding changes. Moreover, Figure~\ref{fig2}\textbf{b)} reveals that, for \emph{Placer}$_{d=2}$, those random seeds that resulted in more dynamic node embeddings also resulted in much lower goodput performance. We conjecture that a possible reason for the quasi-static embeddings is \emph{Placer}'s centralized single-agent inference. It imposes a symmetry on the routing actions, whereas telemetry data may depict asymmetric utilization patterns of links and packet buffers that call for asymmetric routing choices, i.e., different links traversed from $u$ to $z$ than from $z$ to $u$.

\section{Conclusion and Future Work}

\glsresetall

We propose \emph{Placer}, a routing algorithm that uses \glspl{mpn} to turn telemetry-infused network states into latent node embeddings for greedy routing. A fully developed algorithm of this kind could provide sub-second routing responsiveness without relying on shortest-path computations. Additionally, low-dimensional network embeddings in particular could be visualized to help explain the routing decisions made by neural networks. First experiments with \emph{Placer} show that some geometric structure can be extracted, resulting in competitive routing for a five-node network \mbox{topology} example. Yet, more work is needed to fully understand the problem structure, as changes in telemetry data do not yet result in consequential shifts of the node embeddings. This may be a consequence of the current centralized inference design, which may be incapable of providing the asymmetric routing choices required to handle asymmetric network load. Future work may obtain more meaningful node embeddings by deploying \emph{Placer} locally at each node and, perhaps, training it using multi-agent variants of \gls{ppo}~\citep{dewittIndependentLearningAll2020, yuSurprisingEffectivenessPPO2022}. Finally, as many larger networks are more efficiently embedded in hyperbolic spaces~\citep{krioukovGreedyForwardingDynamic2010, bogunaSustainingInternetHyperbolic2010}, we suggest that the Euclidean \gls{mpn} used by \emph{Placer} may be replaced with a hyperbolic variant~\citep{liuHyperbolicGraphNeural2019, chamiHyperbolicGraphConvolutional2019}.

\printbibliography

\end{document}